# A Consumer-tier based Visual-Brain Machine Interface for Augmented Reality Glasses Interactions

**Yuying Jiang[1], Fan Bai[1], Zicheng Zhang[1], Xiaochen Ye[1], Zheng Liu[1], Zhiping Shi[1], Jianwei Yao[1], Xiaojun Liu[1], Fangkun Zhu[1], Junling Li[1], Qian Guo[1], Xiaoan Wang[1], Junwen Luo[1]**

1. BrainUp research lab, BrainUp Co., Ltd, Beijing, China

E-mail :{ jiangyuying; baifan; zhufangkun; zhangzicheng; liuxiaojun; yexiaochen; shizhiping; liuzheng; lijunling; guoqian; yaojianwei; wangxiaoan}@naolubrain.com, junwenluo.ncl.uk@gmail.com

**Abstract** *Objective.* Visual-Brain Machine Interface(V-BMI) has provide a novel interaction technique for Augmented Reality (AR) industries. Several state-of-arts work has demonstates its high accuracy and real-time interaction capbilities. However, most of the studies employ EEGs devices that are rigid and difficult to apply in real-life AR glasseses application sceniraros. Here we develop a consumer-tier Visual-Brain Machine Inteface(V-BMI) system specialized for Augmented Reality(AR) glasses interactions. *Approach.* The developed system consists of a wearable hardware which takes advantages of fast set-up, reliable recording and comfortable wearable experience that specificized for AR glasses applications. Complementing this hardware, we have devised a software framework that facilitates real-time interactions within the system while accommodating a modular configuration to enhance scalability. *Main results.* The developed hardware is only 110g and 120x85x23 mm, which with 1 T ohm and ＜1.5 μVp-p, and a V-BMI based "angry bird" game and an Internet of Thing (IoT) AR applications are deisgned, we demonstrated such technology merits of intuitive experience and efficiency interaction. The real-time interaction accuracy is between 85% and 96% in a commercial AR glasses (DTI is 2.24s and ITR 65 bits/min). *Significance.* Our study indicates the developed system can provide an essential hardware/software framework for consumer based V-BMI AR glasses. Also, we derive several pivotal design factors for a consumer-grade V-BMI-based AR system: 1) Dynamic adaptation of stimulation patterns/classification methods via computer vision algorithms is necessary for AR glasses applications; and 2) Algorithmic localization to foster system stability and latency reduction. These insights lay a foundational groundwork for expediting the advancement of consumer-oriented V-BMI-based AR applications.

## 1. Introduction

Augmented Reality (AR) has emerged as a powerful technique for interacting with the real world, with significant applications in industries such as healthcare[1] and public safety[2]. However, the existing AR interaction technologies, such as gesture and voice control, have limitations that hinder their usability. Gesture control is not ergonomic, as users may experience arm soreness from constantly waving their hands to send commands to an AR system[3]. Voice control[4] is unsuitable in noisy environments and may be socially awkward in public settings.

Also, there is alternative emerging method titled eye tracking is a promising technology for AR communication. Recently there are several states-of-art work demonstrated such technology feasibilities[5–8]. [8]suggested that most participants prefer the gaze-adaptive UI and find it less distracting compared to the others user interfaces, Weighted Pointer[7] interaction enable users to accurately point at targets when eye tracking is accurate and inaccurate which through fallback modalities．However, there are some concerns of these technology applied in practical: (1) limited use in certain conditions: eye-tracking reliability may be affected by changes in lighting conditions, requiring occasional recalibration, incorrect calibration can result in data offset and loss of accuracy. And prolonged eye tracking use may cause eye fatigue, discomfort, or dryness[9]; (2) computational heavily and costly: eye tracking requires significant computational resources to process the data in real-time[10]. This can be a challenge, especially in resource-constrained environments such as mobile devices. In addition, high-quality eye tracking systems can be expensive to develop, implement, and maintain. This can be a barrier for widespread adoption, especially in consumer applications.

Meanwhile, there are a small group of engineers/computational neuroscientists follow another path to develop a novel AR glasses-based interaction technique: translating use's visual attentions into AR digital commands based Visual-Brain Machine Interface(V-BMI)[6,11–21]. For instances, [18]focused on controlling the movement and grasping of a robotic arm using AR technology. [17]combined existing AR headsets with EEG systems to study visual spatial attention in a non-stimulus paradigm, achieving approximately 70% accuracy. [6]also explored the integration of AR and BCI technology, with a potential application in healthcare, but only proposed the idea without actual experiments or data support. [19]integrated AR into a P300-based Brain-Machine Interface (BMI) to control external devices, requiring compatibility between the device and control system to operate. Compare to the eye-tracking technology, visual attention-based paradigm takes advantages of mind-control like experience, light computation load and compact hardware requirements. Although researchers have extensively explored innovative applications of these technologies, there is a consider gap between algorithms[22–25] [26,27]and hardware/system development in this field. This indicates that most studies employed existing EEG devices or relatively rudimentary self-developed devices, these devices are default designed for lab research environment rather than real-world AR glasses application scenarios, and there is no adequate hardware designed for AR glasses. Meanwhile, they lacked a comprehensive software system that could facilitate effective communication of data and commands between AR and V-BMI hardware. These shortcomings hinder this technology towards life scenarios AR glasses-based applications.

To solve this dilemma, we develop a novel AR glasses based Visual-Brain Machine Interface(V-BMI) for real-world applications. As shown in Figure 1, the system consists of two parts: a commercial AR glasses and a specified designed V-BMI hardware. The V-BMI hardware is tailor designed for visual area EEG recordings with seven electrodes which can be fully cover visual cortex areas. Considering AR application in life scenarios, the developed hardware is designed follow principles as below:1) fast setup, 2) reliable signal recording and 3) comfortable user experiences. Meanwhile, a software framework is developed to allow data/commands communications between AR glasses and V-BMI hardware in real-time. The software framework is modular based design that is compatible with the other systems.

The workflow of the system is as follows: first, an AR glasses camera displays current environments on a AR glasses screen. Next, the corresponding interaction menu (stimulation patterns) is generated based on the user's intended task. Meanwhile, V-BMI records the user's electroencephalograms (EEGs) for algorithm processing, which aims to send commands to the AR system directly with the user's visual attention. By requiring only the user's visual attention, the system provides an efficient, private, and immersive AR interaction technology.

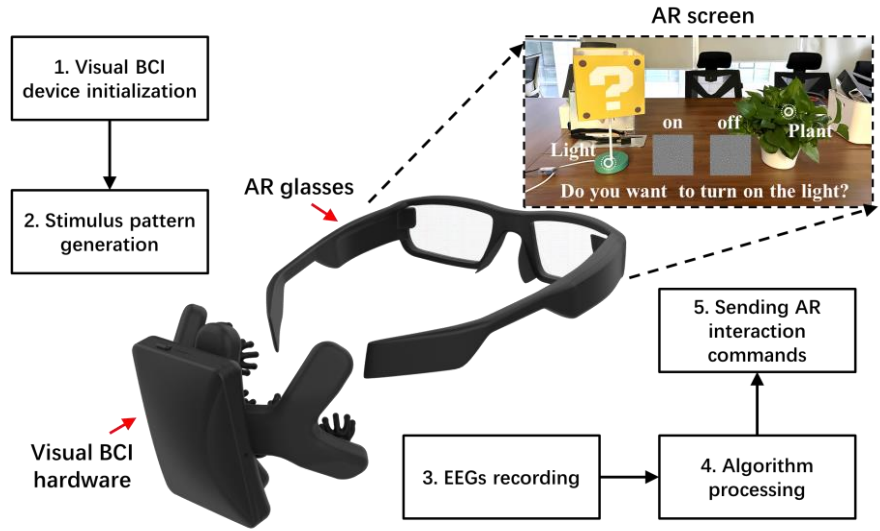

Figure 1: An AR glasses based Visual-Brain Machine Interface (V-BMI) system and a conceptual flow chart. It consists of five steps: visual BCI initializations, stimulus pattern generation, EEG recording, algorithm processing and sending AR interaction commands.

At last, applications of turning on an electrical fan in the physical world via AR hybrid interaction and an Angry bird game was employed to verify the developed system technology feasibility and performances. To summary, the key contributions of this work are as follows:

- Hardware: A V-BMI hardware specificized designed for integrating with an AR glasses. The developed hardware takes advantages of fast setup with dry electrodes, reliable EEG recording with strong artifact-free capabilities and comfortable user wearable experience.
- Software: a modular based software framework is developed for data and commands communication efficiently between AR and V-BMI hardware, this allows V-BMI based AR system can serve as a seamless system for real-time interaction and can be easily custom modified by users' applications.
- Applications: Two novel AR glasses-based V-BMI applications are developed to demonstrate such interaction potential advantages: enhanced immersion and user experience and precise and context-aware actions.

## 2. Methods
### 2.1 AR glasses-based hardware design principles

To meet consumer based AR glasses application requirements, the hardware system is followed FRC (fast setup, reliable recording and comfortable wear) design principles with several design techniques as followed:

(a) Fast setup, the system must be ready for recording reliable EEGs in 2-3 minutes to meet the AR based life scenarios application requirements. A hybrid rigid/flexible multi-layer PCB design in a compact fashion allows the entire system size be 120mm by 85mm, which is easily integrated with an AR glasses. Meanwhile, dry electrodes are employed to reduce the user setup time as well.

(b) Reliable recording, system is frequently used on the life scenarios environments such as office and outdoors. This requests the system has strong noise artifacts-free capabilities. Active electrodes and right leg circuits are employed to serves as this purpose.

(c) Comfortable user wear experience, a human factors engineering designs allow a device can adequate fit human brain shape.

Regarding hardware circuits implementation, it consists of two blocks: analogue recording and digital processing and transmission. The analogue system responses for EEG recordings and can be measured by three factors: input impedance, signal resolutions and anti-environment noise capabilities.

- *Input impedance*

Maintaining the integrity of the obtained signal is imperative as the amplitude of EEG is at the microvolt level, otherwise, the signature signal has the potential to be lost. Figure.2 demonstrates the equivalent circuit of the voltage divider rule. Where $V_{brain}$ is the EEG signal, $V_{out}$ is the remaining EEG signals that can flow into the analog front-end (AFE), $Z$ is the sum equivalent impedance before electrodes and the $Z_{in}$ is the input impedance of the AFE circuit. Specifically, $Z$ consists of $Z_s$ (the skin impedance), $Z_{se}$ (the impedance between the skin and electrode), and $Z_e$ (the impedance of the electrode)[28]. The value of $Z_s$ is vary from 10 Kohm to 1 Mohm and the value of Z_e is approximately 1 Mohm for non-invasive dry electrodes[28,29]. The value of $Z_{se}$ varies dramatically along with the contact situation. Thus, the expected value of $Z$ is around 1Mohm. According to Equation (1), when $Z_{in}$ exceeds $Z$ significantly, the $Z_{in}$ is infinitely close to $V_{brain}$. Consequently, augmenting input impedance is an effective method to guarantee integrity of signals without introducing additional noise.

$$V_{out} = V_{brain} \times \frac{Z_{in}}{Z_{in} + Z} \tag{1}$$

$$loss\ ratio = \frac{Z_{in}}{Z_{in} + Z} \times 100\% \tag{2}$$

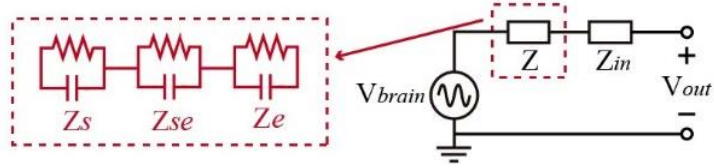

. Figure 2: Impedance equivalent circuit diagram.

- *System signal Resolutions*

$$LSB = \frac{V_{ref}}{2^n} \tag{3}$$

$$V_{min} = \frac{LSB}{gain \times (1 - lossratio)} \tag{4}$$

where, $V_{ref}$ is the reference voltage of ADC, n is the resolution of ADC. System signal resolution means the minimum variation of signals that can be caught by the system, which is mainly determined by the resolution of ADC. The Least Significant Bit (LSB) is the minimum number of digits that can be displayed when an analog signal is converted into a digital signal. According to equations (3) it is defined by ADC's reference voltage and resolution. Furthermore, the gain of the amplifier and loss ratio also have an impact on the system resolution. Equation (4) characterizes the relationship between them. The higher the gain provided; the smaller signals cloud be detected. Conversely, the smaller loss ratio equips the system to collect the smaller signals.

- *Anti-environment noise capability*

$$C_{anti} = \frac{PSD_{50Hz}(V_{out})}{PSD_{50Hz}(V_{in})} \tag{5}$$

The anti-environment noise capability is defined as the capability to eliminate environmental noise[28,29], and to retain the bioelectrical signals at the largest extent, simultaneously. Due to that the largest source of noise in the environment comes from power frequency interference, the anti-environment noise capability is represented by the system's ability to suppress 50Hz. With the useful signals reserved effectively, (5) is utilized to quantify anti-environment noise capability, we summarize it as a ratio of frequency domain value of noise signal(50Hz) at the input-end and output-end of systems after Fourier transform. The smaller ratio of the $C_{anti}$ is, the better anti-environment noise capabilities that systems possess, as 50Hz is suppressed.

## 2.2 Hardware Architecture

The design philosophy of the proposed EEG hardware is constricted by the metrics pointed out above, channels, and portability. Therefore, the design of the system is illustrated in Figure 3(c), it is shaped into a wearable device with 9 channels involving 7 EEG-acquiring, 1 reference[30], and 1 bias. For consideration of high input impedance, a low noise (0.4μV peak to peak) unity-gain buffer which input impedance is greater than 1Tohm is placed beside electrode. According to (2), even if the $Z$ achieves 1Mohm, the signal attenuation is only 0.0001%. Electrostatic Discharge (ESD) aims to prevent static electricity from damaging components. A Resistance Capacitance (RC) Low Pass Filter (LPF) and an RC How Pass Filter (LPF) are placed to eliminate the noise of certain frequencies, the cut-off frequency is 800Hz and 0.3Hz, respectively. An 8-channel synchronous sampling highly integrated Analog to Digital Converter (ADC) is exerted to promote portability, which contains PGA providing the system with an optional gain. Its 24bits signal resolution has potential to catch the signal less than 0.1μV without a gain. Finally, the MCU is the control unit, which communicates with ADC through serial I/O and the workstation through Bluetooth 5.0 protocol.

Notably, to expand the usage scenarios, environmental noise should be eliminated for maximum performance. For this purpose, the Right Leg Driver (RLD) circuit is proposed, the common-mode signals of the acquiring channels extracted at the common-mode ends of the amplifier are fed into the negative input of the inverting-amplifier. Thus, the bias electrode connected to the output of the inverting-amplifier has inverse common-mode signals. In this way, the bias electrode transmits the signals to other electrodes to suppress noise.

The digital system responses for system control and data transmission. The implementation is as below: a hardware processor is based on the ARM Cortex-M4 core. The terminal implements functions such as SD card storage, BLE 5.0 transmission, ADC signal acquisition and amplification, SPI transmission, and program burning debugging. The signal input terminal is for weak brainwave signals. After front-end amplification and filtering, the signals enter the ADC processing unit for mode conversion and amplification. Then, they are transmitted via the SPI bus to the MCU processing unit and output to the server or other processing terminals through Bluetooth 5.0 for signal calculation, restoration, and display. This device has a total of 7 signal acquisition channels and two additional channels for bias and reference to ensure accurate signal acquisition and transmission. The device's power system is powered by a 3.7V lithium battery, which also implements charging functionality, real-time battery monitoring, charging signal indicators, etc.

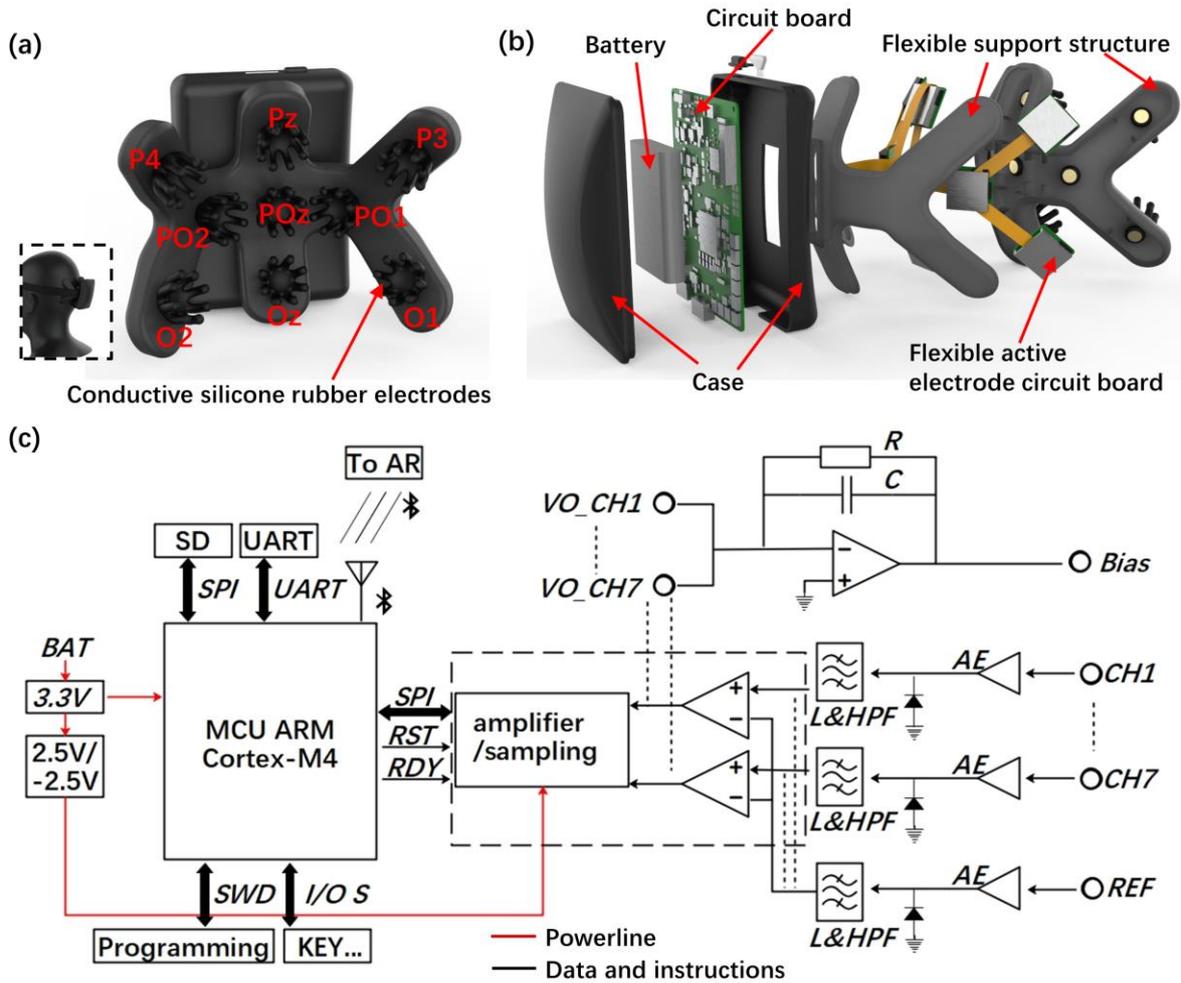

Figure 3: (a) is the hardware concept figure and electrode position diagram. (b) is the diagram of the internal layers of the hardware. (c) is a hardware design schematic.

**2.3 software framework**

To facilitate seamless integration and real-time interaction between AR devices and users, we present a modular based software framework, comprising four distinct layers: an application layer, a SDK layer, a backend service layer, and an algorithm layer. It has two key advantages: 1) Real-time interaction: the framework enables real-time interaction between the user and the AR environment through seamless integration of BCI technology, it incorporates optimized data transmission protocols and rapid processing algorithms, resulting in low latency and fast response times. This ensures that users receive real-time feedback and interactions with minimal delays. 2) Modular and scalable design, new features and technologies as the other researcher developed 0 can be integrated into specific layers without disrupting the overall system functionality, allowing for future expansion and customization.

At the topmost level of the software system, lies the application layer. This layer encompasses two primary components: AR software and IoT smart home applications. The AR software serves as the user interface for AR headsets/glasses, offering intuitive and immersive interactions within the AR environment. Simultaneously, IoT smart home applications empower users to control and monitor various smart home devices through the AR interface. This layer acts as a bridge, connecting the AR device to the underlying software infrastructure.

The Software Development Kit (SDK) layer constitutes the core of the software system, responsible for handling the complexities of BCI integration and visual stimulation paradigms. This layer is further divided into two modules: the EEG service module and the visual stimulation module. The EEG service module plays a vital role in establishing communication with Brain-Computer Interface (BCI) devices and handling EEG data in real-time. Its functions include searching and connecting to BCI devices, receiving and decoding EEG data (reshape EEG data format), buffering and filtering the data (50Hz and 1-100 bandpass filter), and detecting wear states. These operations ensure a seamless flow of EEG data for subsequent processing and analysis. And the visual stimulation module in the SDK layer offers an array of functionalities designed to process V-BMI system. It provides various stimulation paradigms can be selected and displayed on AR screen, algorithm training paradigms and format, and coded Visual Evoked Potential (CVEP) decoding algorithm. Additionally, the module manages data synchronization, thereby guaranteeing the timely presentation of visual stimuli aligned with the user's attention EEG signals.

Beneath the SDK layer resides the backend service layer, which focuses on handling user authentication, processing HTTP requests, and caching user training templates. User authentication ensures secure access to the system, while efficient HTTP request processing ensures smooth communication between the AR headset and the backend servers. Caching user training templates reduces latency during training sessions and improves the overall user experience.

At the core of the software system, the algorithm layer is responsible for EEG data preprocessing, TDCA/TRCA training data templates[16], and inference results. Preprocessing EEG data involves noise reduction and artifact removal to enhance the accuracy and reliability of subsequent processing steps. Deep learning-based algorithms[15] can also be implemented on this layer.

Figure 4(b) is the system software dataflow. The software establishes a BLE connection with the EEG device and sends commands to obtain data. The V-BMI hardware in System-1 is employed to record users' EEG signals at a frequency of 250Hz. Concurrently, an interaction menu is generated as visual stimuli for the user. The captured EEG data is then transmitted to the AR headset via Bluetooth 5.1 at a rate of 13KB/s. Prior to processing, the EEG signals undergo noise filtering, utilizing an algorithm derived from earlier research to enhance signal clarity.

When training begins, the EEG data is synchronized with the stimulation pattern onset, and upon completion, the algorithm generates a user specific training template which stored in the backend service layer. And the TDCA algorithm performs rapid processing within 10ms. At an inference stage, when stimulation starts, the EEG data is synced, and after the stimulation cycle ends, the algorithm is called to obtain results, which are then synchronized and send to AR for a final interaction decision.

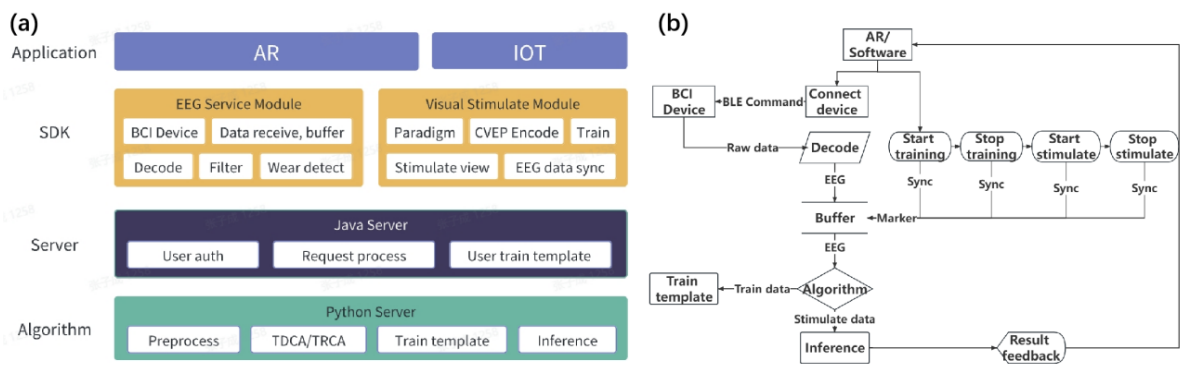

Figure 4: (a) is the system software frame diagram. It consists of four layers: an application layer, a SDK layer, a server and an algorithm layer. (b) is the system software dataflow. It has two process branches: a training and an inference process flow.

## 3. Results
*3.1 Experimental protocol*

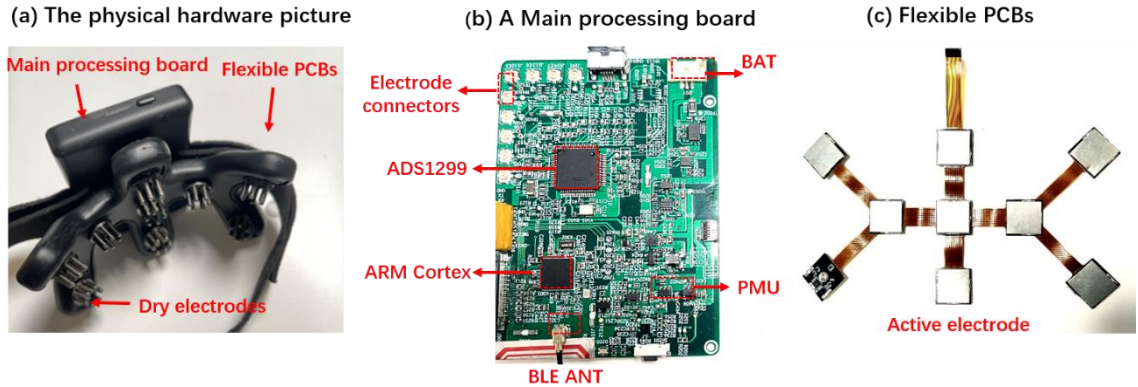

Figure 5: The hardware physical pictures, and it consists of a man processing board and flexible PCBs.

Twenty healthy subjects (18 males and 2females, aged from 22 to 40 years old) participated in the experiments, all of whom had normal or corrected-to-normal visions. All subjects signed the consent form before the experiment. The experiments complied with the Declaration of Helsinki and was approved by the local ethics committee. Wearing both a commercial N-real AR glasses Light (refresh rate 60Hz) and a tailor developed wearable EEG hardware (Figure 5), which locates at the visual cortex areas. subjects sat comfortably in a sound-proof room to complete all the experiments. Before each trial, participants undergo a wear detection process where all indicator lights on the software interface turn green, indicating good connectivity between the electrodes and the scalp, ensuring usable signal quality.

Two AR applications are developed for the system evaluations. The experiment uses a gray checkerboard as the stimulus paradigm primely in an AR glasses, and the cVEP encoding sequence is a 28-bit nearly perfect sequence (1,0,1,1,1,1,1,1,1,1,1,1,1,1,0,0,1,1,1,0,0,0,0,0,0,0,0,0,0,1). The flicker frequency is 25Hz, meaning each trial lasts for 1.12 seconds. For model parameter training, a 30-second training trial is conducted before the actual experiments. The decoding phase consists of four different experiments: 4 targets with 1 trial, 2 trials; and 7 targets with 2 trials, 3 trials. Also, different patterns high-pass filter, pure white, and gray square are sequentially tested in the AR glasses for any bird games.

For EEG, according to the International 10-20 system, the EEG amplifier collected 7-channel EEG data (PO5, PO3, POZ, PO4, O1, OZ, O2), and Cz and AFz were reference and ground electrodes, respectively. The sampling rate was set to 250 Hz. The impedance of each channel was kept less than 50 kΩ. The data and command communications via a software framework.

*3.1 Turning on an electrical fan in the physical world via AR hybrid interaction*

To explore proposed technology potentials, we employ a scenario where a user is wearing a V-BMI combined AR glasses in a smart home environment. The physical smart home environment has a digital twin on AR glasses (Figure 6b and 6c). Therefore, by combining hand gestures and visual attention control, the user can rotate digital world area by moving hands while direct their visual attentions towards an electrical fan stimulus pattern in the AR glasses, which triggers the command to turn on the corresponding fan in the physical world (Figure 6a). As it shown in Figure 5d, a user wears a V-BMI and AR glasses, by utilizing spatial virtual modelling, the AR glasses display the 3D model and current status of the designated physical space (Figure 6e). Users can rotate or zoom in on the space, they select the space they want to remotely control by hands and use visual brain-computer interaction to control the switches and other states of the appliances, achieving real-time remote control, the physical electrical fan was switched on in the AR glasses as an example (Figure 6f). This demonstrates the seamless integration of gesture and visual attention control for real-time manipulation of physical objects using AR technology. The full video of this application is at Supplement A.

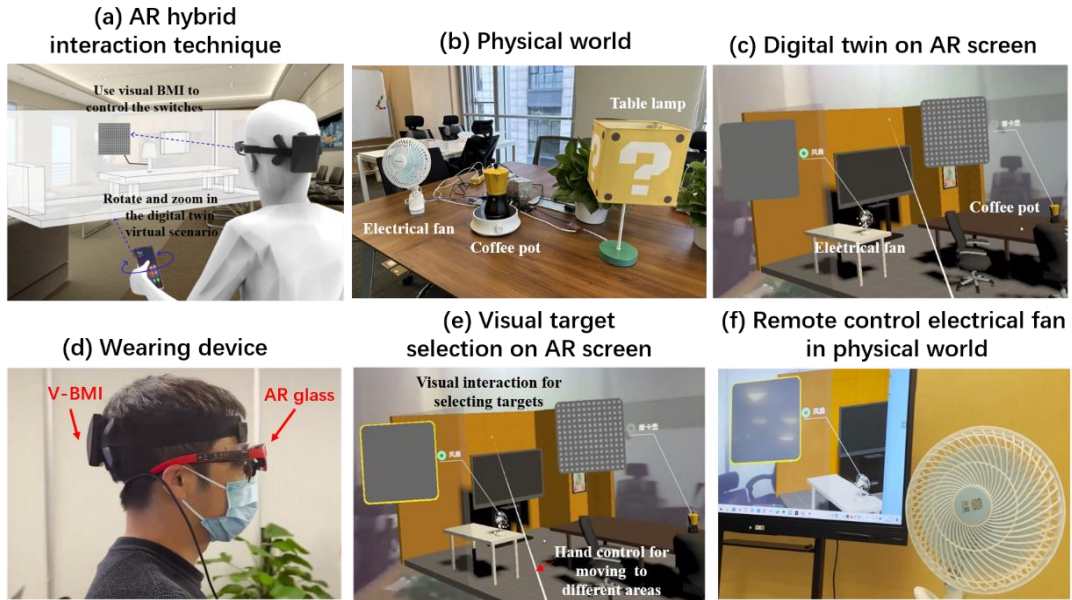

Figure 6: An Internet of Things (IoT) application demo. (a): a AR hybrid interaction technique that use visual attentions to control the switches, and use hand gesture to rotate and zoom in the digital world simultaneous; (b): a physical world with an electrical fan, a table lamp and a coffee pot;(c) a digital twin of a physical world on AR screen; (d) a picture of a user wearing both a V-BMI and a AR glasses; (e) the visual target selection on AR glasses screen and (f): a remote control electrical fan in physical world.

## 3.3 V-BMI based "Angry Birds" game

Figure 7 is a demo of the game "Angry Birds". This demo transforms the interaction of dragging the slingshot to shoot with a finger into a visual brain-computer interaction to select the target to attack, integrating visual brain-computer interaction technology into the mixed reality (MR) game to provide users with a multimodal gaming experience. Users wear AR glasses and the V-BMI device, and after completing the pre-training (Figure 7a), they start the game. In the game, specific 3D blocks where the target birds are located are covered with stimulating materials that can blink according to a specified *cVEP* encoding method. As the user moves in the 3D space, the stimulating materials adjust their texture scale based on the user's position, ensuring the spatial frequency stability of the stimulating materials and ensuring accurate recognition (Figure 7b). Users can gaze at the intended block to select it, triggering the interactive attack in the game to knock down the target bird and score to progress to the next level (Figure 7c). The full video of this application is at Supplement B.

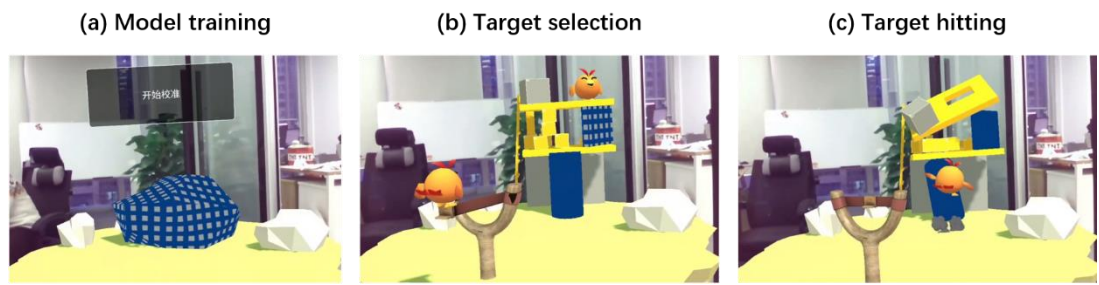

Figure 7: a demo of the game "Angry Birds".

## 3.2 System accuracy on computer screen

The results, shown in Figure 8, indicate that the accuracy for the 3-trial 7-target experiment is 96.11%, with a standard error of ±1.52%; the accuracy for the 2-trial 7-target experiment is 92.22%, with a standard error of ±4.55%; the accuracy for the 2-trial 4-target experiment is 96.53%, with a standard error of ±2.49%; and the accuracy for the 1-trial 4-target experiment is 92.36%, with a standard error of ±1.92%. It can be observed that increasing the number of trials is more likely to improve data accuracy and reliability. All results show accuracy above 92%, demonstrating that the device, algorithm, and experimental paradigm are robust and applicable in the field of V-BMI.

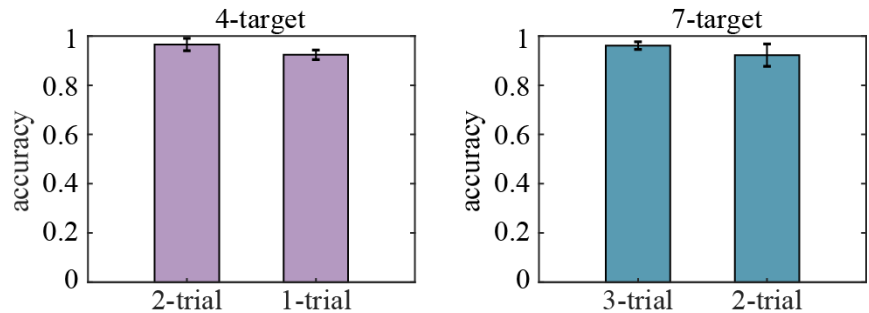

Figure 8: system accuracy on computer screen.

### 3.3 System accuracy on AR glasses

We used the "Angry Birds" demo to test the accuracy on AR screen. We tested three different stimulus patterns: high-pass filter, pure white, and gray square (Figure 9(a)). In the "Angry Birds" demo, there are different levels that correspond to different levels of environment complexity. In the simple environment, there are only 2-3 goals, in the complex environment, there are 5-6 goals.

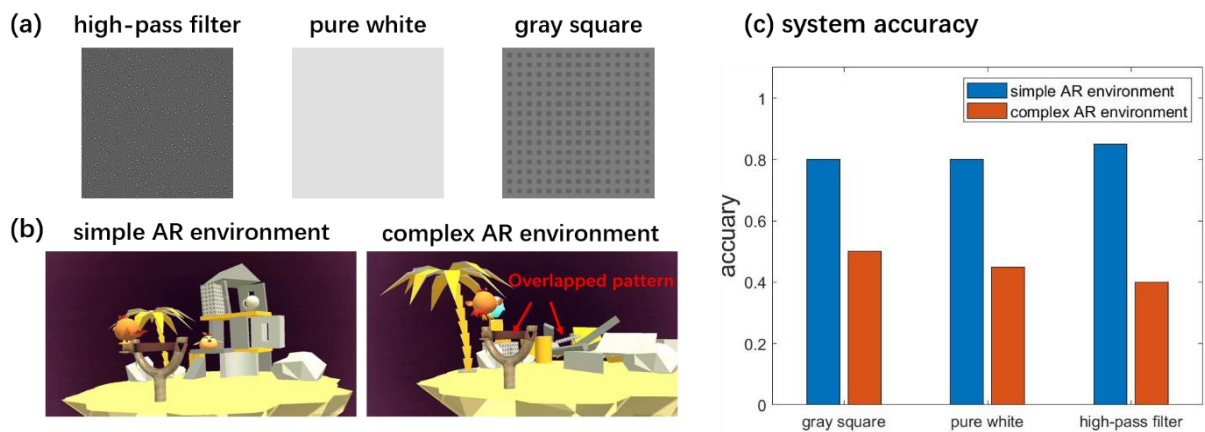

Figure 9: patterns display accuracy on AR screen. (a) are the different patterns of stimulation. (b) is the picture of simple and complex environment, the red arrows are the targets.

The accuracy was shown in the figure 9c. In the simple environment, the accuracy of the gray square and pure white patterns is 80%, and the accuracy of the high-pass filter pattern is 85%. This is close to the accuracy on the computer screen (96%). This indicates that too much brightness on a stimulation pattern in AR glasses may decrease system accuracy. In the complex environment, the accuracy of the gray square is 50%, the accuracy of the pure white pattern is 45%, and the accuracy of the high-pass filter pattern is 40%. As can be seen, the accuracy will decrease in complex environment. The results are identical with[14]. Mainly because the number of targets in the complex environment has increased, and with the progress of the game, there will be overlap or occlusion between the targets, which causes a large degree of mutual interference between the targets. Therefore, in actual use, try to build a simple scene, or make the target more dispersed, which can avoid a significant reduction in accuracy to a certain extent.

### 3.4 AR based V-BMI comparisons

Three other state-of-the-art Augmented Reality (AR) based Visual Brain-Machine Interface (V-BMI) systems have been utilized for the purpose of comparative analysis. From a hardware standpoint, our system capitalizes on its compact form factor(120x85x23mm and 110g) compatible with AR glasseses and its elevated input impedance, necessitating only a minimal seven-channel configuration. Conversely, it should be noted that [18]and [19]exhibit a substantial hardware footprint, rendering them potentially impractical for real-world deployment. Additionally, [18]demonstrates a comparatively reduced input impedance of 47 GΩ, which could potentially compromise signal fidelity within outdoor settings. Although [17] boasts a commendable hardware design conducive to AR headset integration, it regrettably lacks in-depth quantitative assessments of hardware performance metrics.

Regarding the software and algorithmic dimensions, our investigation has yielded accuracy levels exceeding 85% when implemented on AR glasses for IoT control and gaming applications. This achievement stands in contrast to the findings of [17], where the application was executed on a Hololens glasses. In a distinct vein, the proposed covert visuospatial attention (CVSA) paradigm introduced by [17]eliminates the necessity for externally induced stimulus-driven responses, potentially augmenting user engagement. However, the empirical substantiation of its efficacy necessitates dedicated application. Conversely, [18,19]and [19] demonstrate comparable performance outcomes, upon computer screens or LCDs as display mediums. Integration of machine vision allows the stimulation pattern dynamically modified in a manner. However, the non-wearable nature of these setups imparts a heightened stability to the stimulus generation process, consequently mitigating certain algorithmic intricacies and signal-to-noise ratio requisites inherent to wearable configurations, and translate these stimulation patterns to a AR glasses may cause system accuracy drops.

Table 1: The comparison of AR based V-BMI system performances.

| | This work | [18] | [19] | [17] |
|---|---|---|---|---|
| Hardware specifications | | | | |
| Channel count | 7 | 32 | 16 | 14 |
| Size(mm) | 120x85x23 | | 197x155x40 | |
| Weight (g) | 110 | - | 1000 | 270 |
| Wearable/wireless | Y/Y | Y/Y | Y/N | Y/Y |
| ADC resolution (bit) | 24 | 16 | 24 | 24 |
| Sampling rate (SPS) | 250 | 300 | >128 | 250 |
| Noise level (Shorted inputs) | < 1.5 $\mu V_{p-p}$, < 0.5 $\mu V_{RMS}$ 1-125 Hz | <3$\mu V_{p-p}$ | < 1.5 $\mu V_{p-p}$, 1-30Hz | - |
| Input impedance | >1 TΩ | 47 GΩ | > 1000 GΩ ( 220 pF) | - |
| Active electrodes | Y | - | Y | - |
| CMRR @ 50 Hz | > 110db | >120db | - | - |
| Software specifications | | | | |
| Algorithms | TDCA | Modified FBCCA | Feature vector | |
| ITR | 65 | 56.4 | - | |
| DTI | 2.24s | depends | 2.25s | 3s |
| Target number | 7 | 7 | 4 and 11 | 2 |
| Accuracy (%) | 85-96 | depends | 82.7-88 | 70 |
| Applications | IoT control and gaming | Robot arm control | IoT control | - |
| Pattern Display | AR glasses | Computer screen | LCD/HMD | AR glasses |
| Stimulation pattern | cVEP | SSVEP | P300 | CVSA |

*3.5 Strong anti-artefact capabilities*

The V-BMI device is designed to seamlessly adapt to various working environments. Users can conveniently transmit and store data through SD card or wireless Bluetooth or USB data communication, allowing the system to operate in different conditions such as the office, home, meeting room or outdoors (Figure 10(a, b)).

The V-BMI device also establishes a good foundation for the extraction of pure EEG signals with portable structures. Foremost, as shown in Figure 9, the system peak-to-peak value of input noise is 1μV. Namely, our noise level is less than 0.5μVrms at1-125 Hz, which means that the EEG signals greater than 0.5μV can be extracted by our device, which can fully satisfy the algorithm requirement. Accordingly, our Common Mode Rejection Ratio (CMRR) is 110dB, which makes the common mode signal (noise) can be well eliminated. Second, our input impedance is more than 1Tohm, which is far greater than the contact impedance between the electrode and the skin. Consequently, our device has a good effect on guaranteeing signal integrity.

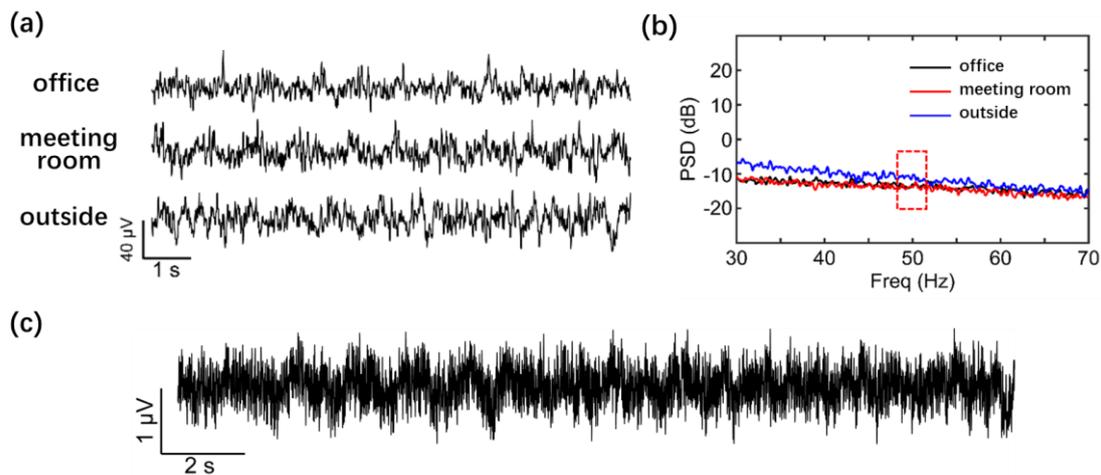

Figure 10: The V-BMI device has strong anti-interference capability. (a) is the original signal waveform of EEG in different environments. (b) is the PSD of EEG in different environments. (c) is the system noise level.

## 4. Discussion

*4.1 The advantages of AR based hybrid V-BM interaction*

V-BMI based AR interaction systems hold several potential advantages that contribute to a more enriched user experience. These benefits span various dimensions, starting with the facilitation of natural and intuitive interactions. By seamlessly merging gesture control with visual attention control, users are empowered to engage with augmented reality (AR) elements in a manner that feels innate and intuitive. Employing hand gestures to initiate commands and concurrently directing their visual focus toward specific objects or Internet of Things (IoT) devices, users effortlessly bridge the gap between the physical and virtual domains.

Moreover, this integration of gesture and visual attention control yields precise and context-aware actions within the AR environment. This amalgamation empowers users to harness their intrinsic spatial awareness and motor skills, translating into the execution of highly targeted commands. Consequently, the synergy between these modalities enhances the accuracy and efficiency of interactions with both virtual and real-world entities, fostering a more dynamic and effective AR experience.

*4.2 Design factors for AR based V-BM system*

The realization of an AR based V-BMI necessitates a comprehensive hardware and software co-design strategy that aligns with the system's fundamental prerequisites: 1) the intricate nature of typical V-BMI algorithms warrants their implementation to be primarily localized (e.g., employing languages like C++) as opposed to cloud-based approaches. This strategic localization diminishes system latency by mitigating data transfer delays between the SDK hardware and remote servers, thereby enhancing system stability. There already established several works for hardware/software co-design [31,32] that developed low-lower accelerator for implementing several key algorithms. 2) the sampling frequency for EEG recordings must harmonize with the data transmission rate of Bluetooth communication to avert data overflow or substantial latency. Consequently, we have opted for a recording frequency of 250Hz, enabling the Bluetooth module to transmit 250 data points per second.3) it is imperative to acknowledge that algorithmic accuracy may attenuate under complex AR conditions (Figure 9)[16][16]contributed online adaptive CCA algorithm to improve the performances of the AR-BCI applications in complex lighting environments. Instances include situations where two closely-spaced stimulus patterns coexist or when patterns diminish in size significantly. Addressing this, here we suggest a computer vision algorithm emerges as a necessary tool to dynamically amplify and enhance stimulus patterns contingent or modify pattern brightness on prevailing environmental factors.

**Authorship**
Yuying Jiang was responsible for the execution of application testing and precision assessments for the purpose of evaluating system accuracy. Fai Bai and Zhiping Shi were involved in the formulation of hardware design strategies and their subsequent implementations. Qian Guo, Xiaochen Ye, and Junling Li played instrumental roles in the progression of Unity SDK IoT and game application development. Zicheng Zhang and Junling Li contributed to the advancement of software framework establishment. Zheng Liu was dedicated to the refinement of algorithmic localization techniques and the enhancement of system latency. Xiaojun Liu and Jianwei Yao undertook the pivotal task of conducting comprehensive hardware signal analyses and performance evaluations. FangKun Zhu took charge of algorithmic formulation and refinement. Junwen Luo assumed the role of the principal investigator overseeing the entirety of the project.

**Reference**

[1]     Hsieh M C and Lee J J 2018 Preliminary Study of VR and AR Applications in Medical and Healthcare Education *Journal of Nursing and Health Studies* **03**

[2]     Li X, Yi W, Chi H L, Wang X and Chan A P C 2018 A critical review of virtual and augmented reality (VR/AR) applications in construction safety *Autom Constr* **86**

[3]     Liang H, Yuan J, Thalmann D and Nadia M T 2015 AR in hand: Egocentric palm pose tracking and gesture recognition for augmented reality applications *MM 2015 - Proceedings of the 2015 ACM Multimedia Conference* 743–4

[4]     Sheldon A, Dobbs T, Fabbri A, Gardner N, Haeusler H, Ramos C and Zavoleas Y 2022 Putting the AR in (AR)chitecture - Integrating voice recognition and gesture control for Augmented Reality interaction to enhance design practice *Proceedings of the 24th Conference on Computer Aided Architectural Design Research in Asia (CAADRIA)* vol 1

[5]     Elmadjian C, Shukla P, Tula A D and Morimoto C H 2018 3D gaze estimation in the scene volume with a head-mounted eye tracker *Proceedings - COGAIN 2018: Communication by Gaze Interaction*

[6]     Ha J, Park S and Im C-H 2022 Novel Hybrid Brain-Computer Interface for Virtual Reality Applications Using Steady-State Visual-Evoked Potential-Based Brain–Computer Interface and Electrooculogram-Based Eye Tracking for Increased Information Transfer Rate *Front Neuroinform* **16**

[7]     Sidenmark L, Parent M, Wu C H, Chan J, Glueck M, Wigdor D, Grossman T and Giordano M 2022 Weighted Pointer: Error-aware Gaze-based Interaction through Fallback Modalities *IEEE Trans Vis Comput Graph* **28**

[8]     Piening R, Pfeuffer K, Esteves A, Mittermeier T, Prange S, Schröder P and Alt F 2021 Looking for Info: Evaluation of Gaze Based Information Retrieval in Augmented Reality *Lecture Notes in Computer Science (including subseries Lecture Notes in Artificial Intelligence and Lecture Notes in Bioinformatics)* vol 12932 LNCS

[9] Burch M, Haymoz R and Lindau S 2022 The Benefits and Drawbacks of Eye Tracking for Improving Educational Systems *Eye Tracking Research and Applications Symposium (ETRA)*

[10] Jansen A R, Blackwell A F and Marriott K 2003 A tool for tracking visual attention: The Restricted Focus Viewer *Behavior Research Methods, Instruments, and Computers* **35**

[11] Du Y and Zhao X 2022 Visual stimulus color effect on SSVEP-BCI in augmented reality *Biomed Signal Process Control* **78**

[12] Zhao X, Liu C, Xu Z, Zhang L and Zhang R 2020 SSVEP Stimulus Layout Effect on Accuracy of Brain-Computer Interfaces in Augmented Reality Glasses *IEEE Access* **8**

[13] Fang B, Ding W, Sun F, Shan J, Wang X, Wang C and Zhang X 2022 Brain-Computer Interface Integrated with Augmented Reality for Human-Robot Interaction *IEEE Trans Cogn Dev Syst*

[14] Zhang R, Xu Z, Zhang L, Cao L, Hu Y, Lu B, Shi L, Yao D and Zhao X 2022 The effect of stimulus number on the recognition accuracy and information transfer rate of SSVEP-BCI in augmented reality *J Neural Eng* **19**

[15] Zhao X, Du Y and Zhang R 2022 A CNN-based multi-target fast classification method for AR-SSVEP *Comput Biol Med* **141**

[16] Zhang R, Cao L, Xu Z, Zhang Y, Zhang L, Hu Y, Chen M and Yao D 2023 Improving AR-SSVEP Recognition Accuracy Under High Ambient Brightness Through Iterative Learning *IEEE Transactions on Neural Systems and Rehabilitation Engineering* **31**

[17] Kosmyna N, Hu C Y, Wang Y, Wu Q, Scheirer C and Maes P 2020 A Pilot Study using Covert Visuospatial Attention as an EEG-based Brain Computer Interface to Enhance AR Interaction *Proceedings - International Symposium on Wearable Computers, ISWC*

[18] Zhang D, Liu S, Wang K, Zhang J, Chen D, Zhang Y, Nie L, Yang J, Shinntarou F, Wu J and Yan T 2021 Machine-vision fused brain machine interface based on dynamic augmented reality visual stimulation *J Neural Eng* **18**

[19] Takano K, Hata N and Kansaku K 2011 Towards intelligent environments: An augmented reality-brain-machine interface operated with a see-through head-mount display *Front Neurosci*

[20] Zhu F, Zhang Z, Chen S S, Guo Q, Ye X, Liu X, Wang X, Lin C and Luo J 2022 Live Demonstration: A Stimulation Senseless based Visual Brain-computer Interface for Shopping Scenarios *BioCAS 2022 - IEEE Biomedical Circuits and Systems Conference: Intelligent Biomedical Systems for a Better Future, Proceedings* 246

[21] Chen X, Zhao B, Wang Y and Gao X 2019 Combination of high-frequency SSVEP-based BCI and computer vision for controlling a robotic arm *J Neural Eng* **16**

[22] Liu B, Chen X, Shi N, Wang Y, Gao S and Gao X 2021 Improving the Performance of Individually Calibrated SSVEP-BCI by Task- Discriminant Component Analysis *IEEE Transactions on Neural Systems and Rehabilitation Engineering* **29**

[23] Chen X, Wang Y, Gao S, Jung T P and Gao X 2015 Filter bank canonical correlation analysis for implementing a high-speed SSVEP-based brain-computer interface *J Neural Eng* **12**

[24] Chen Y, Yang C, Chen X, Wang Y and Gao X 2021 A novel training-free recognition method for SSVEP-based BCIs using dynamic window strategy *J Neural Eng* **18**

[25] Lotte F, Bougrain L, Cichocki A, Clerc M, Congedo M, Rakotomamonjy A and Yger F 2018 A review of classification algorithms for EEG-based brain-computer interfaces: a 10 year update *J Neural Eng* **15**

[26] Jiang J, Yin E, Wang C, Xu M and Ming D 2018 Incorporation of dynamic stopping strategy into the high-speed SSVEP-based BCIs *J Neural Eng* **15**

[27]     Zhang S, Han X, Chen X, Wang Y, Gao S and Gao X 2018 A study on dynamic model of steady-state visual evoked potentials *J Neural Eng* **15**

[28]     Habibzadeh Tonekabony Shad E, Molinas M and Ytterdal T 2020 Impedance and Noise of Passive and Active Dry EEG Electrodes: A Review *IEEE Sens J* **20**

[29]     Zhang C, Sabor N, Luo J, Pu Y, Wang G and Lian Y 2021 Automatic Removal of Multiple Artifacts for Single-Channel EEG *Journal of Shanghai Jiaotong University (Science) 2021* 1–15

[30]     Hu S, Lai Y, Valdes-Sosa P A, Bringas-Vega M L and Yao D 2018 How do reference montage and electrodes setup affect the measured scalp EEG potentials? *J Neural Eng* **15**

[31]     Lyu S, Chowdhury M H and Cheung R C C 2023 Efficient Hardware and Software Co-design for EEG Signal Classification based on Extreme Learning Machine *3rd International Conference on Electrical, Computer and Communication Engineering, ECCE 2023*

[32]     Sriram K, Karageorgos I, Wen X, Vesely J, Lindsay N, Wu M, Khazan L, Pothukuchi R P, Manohar R and Bhattacharjee A 2023 HALO: A Hardware-Software Co-Designed Processor for Brain-Computer Interfaces *IEEE Micro* **43**